\documentclass[letterpaper, 10 pt, journal, twoside]{ieeetran}
\IEEEoverridecommandlockouts 
\usepackage{graphicx}

\usepackage{amsmath,amssymb}
\usepackage{amsxtra}

\usepackage{xcolor}
\definecolor{blue}{rgb}{0,0,0}
\DeclareMathOperator*{\argmin}{arg\,min}
\title{Human-in-the-loop Auditory Cueing Strategy for Gait Modification}

\author{Tina LY Wu$^{1}$, Anna Murphy$^{2}$, Chao Chen$^{3}$, and Dana Kuli{\'c}$^{4}$ 

\thanks{$^{1}$First author is with Department of Electrical and Computer Systems Engineering, Monash University, Australia
    {\tt\footnotesize lee.wu@monash.edu}}%
\thanks{$^{2}$Second author is with Clinical Research Centre for Movement Disorders \& Gait, Monash Health, Australia 
    {\tt\footnotesize anna.murphy2@monashhealth.org}}%
\thanks{$^{3}$Third author is with Department of Mechanical and Aerospace Engineering, Monash University, Australia
    {\tt\footnotesize chao.chen@monash.edu}}%
\thanks{$^{4}$Fourth author is with Department of Electrical and Computer Systems Engineering and Department of Mechanical and Aerospace Engineering, Monash University, Australia
    {\tt\footnotesize dana.kulic@monash.edu}}%
}
\begin{document}



\maketitle
\begin{abstract}
External feedback in the form of visual, auditory and tactile cues has been used to assist patients to overcome mobility challenges. However, these cues can become less effective over time. There is limited research on adapting cues to account for inter and intra-personal variations in cue responsiveness. We propose a cue-provision framework that consists of a gait performance monitoring algorithm and an adaptive cueing strategy to improve gait performance. The proposed approach learns a model of the person's response to cues using Gaussian Process regression. The model is then used within an on-line optimization algorithm to generate cues to improve gait performance. We conduct a study with healthy participants to evaluate the ability of the adaptive cueing strategy to influence human gait, and compare its effectiveness to two other cueing approaches: the standard fixed cue approach and a proportional cue approach. The results show that adaptive cueing is more effective in changing the person's gait state once the response model is learned compared to the other methods. 
\end{abstract}

\begin{IEEEkeywords}
Human Factors and Human-in-the-Loop, Rehabilitation Robotics, Wearable Robotics
\end{IEEEkeywords}


%
\IEEEpeerreviewmaketitle

\section{Introduction}
\IEEEPARstart{A}{ssistive} robots have been applied in gait rehabilitation for Parkinson's Disease (PD), stroke, spinal cord injury, and others \cite{lunenburger2007biofeedback, louie2016powered}. The robots can take the form of exoskeletons (e.g. Lokomat and ReWalk \cite{louie2016powered}), capable of monitoring the patients' joint motion in real time and providing assistive torques to guide the movement of patients \textcolor{blue}{who require weight bearing assistance} \cite{lunenburger2007biofeedback}. \textcolor{blue}{On the other hand, patients capable of weight bearing can use wearable devices that provide simple feedback such as visual, auditory, and tactile cues to help their gait in both rehabilitation and everyday settings} \cite{lunenburger2007biofeedback, sweeney2019technological}. Visual cues such as laser projections provide spatial information on where to step, whereas auditory and tactile cues provide temporal information on when to step using metronome beats or vibrations \cite{sweeney2019technological}. 


Existing research on cues focuses on providing visual cues at a fixed distance or auditory/tactile cues at a fixed pace calibrated to each patient \cite{delval2008effect, schaefer2014auditory, Bachlin2010, Mikos2019}. \textcolor{blue}{These cue provision paradigms have several limitations over long-term use and for diseases with progressive symptoms such as PD}. For instance, the cueing mechanisms are often used in conjunction with medications for PD treatment, and the same patient can respond differently to the cues depending on the medication state \cite{sweeney2019technological}. In addition, long-term use of the cues can result in habituation, where the cues become less salient and lose their effectiveness. Patients \textcolor{blue}{might also become reliant on the cues} even without gait abnormalities \cite{ginis2018cueing}. Current cueing mechanisms do not address symptom fluctuation, habituation or cue-dependency and thus, there is a need to develop a cue adaptation strategy to address issues with static cues. 

We propose an adaptive cue-provision framework that can simultaneously monitor the person's gait performance and provide personalized cues to change the person's gait to a \textcolor{blue}{target} state. Personalized cues are provided by continuously learning a model of the individual's response to the provided cues, and utilizing the model to optimise cue selection. The performance of the adaptive cueing strategy is compared to two alternative approaches, the fixed cue and the proportional cue. The fixed approach implements the typical cueing approach in the literature. The proportional approach is a semi-adaptive strategy that generates cues based on individual user performance, but using a fixed control strategy. The results show that adaptive cueing outperforms the other two cueing methods in changing the participant's gait once the personalized response model has been learned. 

\section{Related Work}\label{sec:related_work}
\textcolor{blue}{A review of the} two primary features in an assistive feedback system, patient monitoring and providing personalized feedback, is presented in this section.

\subsection{Monitoring Gait Parameters}
Gait performance can be quantified using parameters such as stride length, cadence, velocity, and double support time \cite{conklyn2010home}. Methods for calculating gait parameters for cue-provision can be categorized into approaches that are used in clinical or everyday settings. 

\textcolor{blue}{In clinical settings, commercially available marker (e.g. VICON Motion System) or pressure-mat (e.g. GAITRite System) measuring devices can be used \cite{conklyn2010home, wu2020novel}. Marker-based systems use cameras to track markers attached on the limbs, from which angles are derived to compute gait parameters. Pressure-mat based systems estimate the parameters using the position and timing of the foot landing on a pressure-sensitive mat. While these systems are considered the gold standard, they require specialized equipment and hence are typically employed in validation studies, rather than being used as a feedback signal to a cueing system.} 

Sensors that are portable, unintrusive, and easy to set up, such as inertial measurement units (IMU) or encoders, have been embedded onto wearable devices to measure gait metrics outside of the clinical setting and provide information for \textcolor{blue}{cue adjustment}. For instance, in \cite{wu2020novel}, the authors developed a laser projection system that can be mounted onto a walking frame. The system adjusts the location of the visual cues based on the movement of the person measured through encoders embedded on the walking aid to ensure that the projection is always a fixed distance ahead of the person. Stride length has also been measured through sensor fusion algorithms using the gyroscope and accelerometer signals from IMUs to adjust for the location of the visual cues \cite{ahn2017smart}.

\subsection{Providing Personalized Assistance}
\textcolor{blue}{Personalization of the cues usually involves changing cue modality, location, and form factors. For instance, visual cue personalization can include adjusting the projection according to the user's step length, or changing the location of the projection device on the user} (e.g. foot-strapped wearables like Path Finder LaserShoes (Walk with Path, Essex, England), walking aid based system like U-Step Walker (U-Step Mobility Products, Inc; Illinois, USA), or Augmented-reality glasses \cite{ahn2017smart}). 

Auditory and tactile cue adjustments share common tuning parameters, such as the duration of the cues (i.e. continuous or on-demand), frequency of the cues (i.e. set speed or patient-specific speed), and timing of the cues (e.g. reactive or proactive, synchronization to the gait cycle events). \textcolor{blue}{A characteristic unique to auditory cues is the provision of music melody, human voice, or metronome beats \cite{Ghai2018}}. The parameters unique to tactile cues are the amplitude and the pattern of the cue. Various forms of vibration have been tested (e.g. constant \cite{Punin2018}, or variable \cite{yasuda2020development}) and can be provided via electrical stimulation \cite{rosenthal2018sensory} or vibration motor \cite{Punin2018}. The effect of the tactile cue location has also been examined \cite{ sweeney2019technological, rosenthal2018sensory, pereira2016freezing}. \textcolor{blue}{A recent study developed a wearable system that provides tactile cues to induce a new walking speed in healthy participants \cite{zhang2020wearable}. The system, which consists of pressure resistive sensors, ERM motors, and an IMU sensor, monitors walking speed and gait events and adjusts feedback provided by the motors using a PI controller. While the closed-loop system was better at inducing the desired speed changes compared to an open-loop algorithm that provided feedback at a constant pace, it is unclear how the controller gains were selected. As well, the same gains were used across all participants.} 

Online feedback adaption, including human-in-the-loop (HIL) optimization, has been investigated for robotic exoskeletons. In the HIL framework, real-time adjustment of the assistance is implemented based on the current performance of the user. Specifically, the HIL framework has been applied in optimizing the assistive force provided by exoskeletons to reduce metabolic cost during walking \cite{kim2017human, zhang2017human, felt2015body}. A fundamental requirement for HIL is building a model that relates the input assistive force to the output performance metrics. Previous studies have used a set of pre-defined assistive forces to uniformly explore the parameter space for the model \cite{zhang2017human}. Others have also investigated more sample-efficient methods without the initial parameter exploration by using gradient descent \cite{kim2017human, felt2015body} or Bayesian optimization \cite{kim2017human}. \textcolor{blue}{Overall, the current cue adaptation strategy in gait rehabilitation is limited to the one-time adjustment to calibrate the cue for the user's height, preferred cadence, or preferred location. While adaptive cue-provision is under investigation, there is a lack of personalization to account for the user's immediate motor capability and response}. 

\section{Proposed Approach}
\label{sec:propApp}
\begin{figure*}[htb]
\begin{center}
\includegraphics[width=\textwidth]{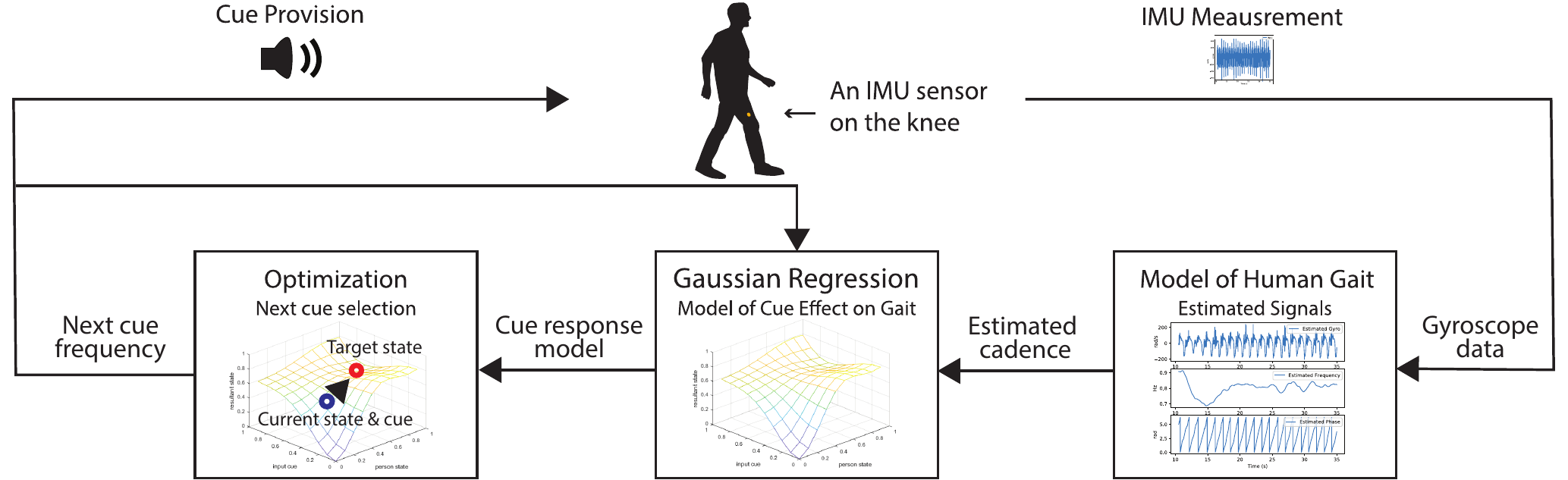}
\caption{\textcolor{blue}{The proposed system is a feedback loop that consists of the human, gait measurement and estimation, cue response learning and cue provision. The human gait model computes metrics to monitor gait performance using the data from the IMU. The Gaussian process (GP) regression model then uses the gait performance metrics, along with the history of the provided cues, to model the gait performance as a function of the provided cues. Finally, the optimization algorithm utilizes the GP model to provide personalized cues that would prompt the participant's current state (the blue circle in the optimization block) to move towards the target state (the red circle)}.}
\label{sysID}
\end{center}
\end{figure*}

The proposed adaptive cue-provision framework, shown in Figure~\ref{sysID}, can continuously monitor the person's gait performance and periodically adjust the assistance based on the person's response to the feedback. 

\subsection{Online Gait Parameter Estimation}
The canonical dynamical system (CDS) proposed in \cite{petrivc2011line} \textcolor{blue}{is used in the study. The system models periodic signals using Fourier series and} has been previously applied in online learning and modelling of an individual's gait \cite{waugh2019online}. The gait is captured by a single inertial measurement unit (IMU) fixed above the individual's knee of the dominant leg. The sensor is oriented such that the y-axis aligns with the normal of the sagittal plane. Once the gait model is learned, the associated model coefficients allows metrics to be derived for continuous monitoring. The CDS is defined as:

\begin{equation}
    \hat{y_t} = \sum_{m=0}^{M}\hat\alpha_{m,t} \sin(m\hat\phi_t) + \hat\beta_{m,t} \cos(m\hat\phi_t) 
    \label{eq:cds}
\end{equation}

\noindent where $\hat{y_t}$ is the estimated signal, $t$ is the current timestep, \textit{M} is the total number of harmonics, $\hat\alpha_{m,t}$ and $\hat\beta_{m,t}$ are the Fourier series coefficients associated with the $m^{th}$ harmonic, and $\hat\phi_t$ is the phase of the signal. The coefficients are updated iteratively through the equations below:
\begin{align}
&e_t=y_t-\hat y_t \notag \\
&\hat\phi_{t+1}= mod(\hat\phi_t+T(\hat\omega_t-\mu e_t sin(\hat\phi_t)),2\pi) \notag \\
&\hat\omega_{t+1}= \mid \hat\omega_{t}-T\mu e_t sin(\hat\phi_t)\mid \notag \\
&\hat\alpha_{m,t+1}=\hat\alpha_{m,t}+T\eta e_t cos(m\hat\phi_t)\notag \\
&\hat\beta_{m,t+1}=\hat\beta_{m,t}+T\eta e_t sin(m\hat\phi_t) \notag 
\end{align}

\noindent where $y$ is the input signal to be learned (i.e. the gyroscope signal in the y-axis), $\hat\omega$ is the estimated frequency, $T$ is the sampling period in seconds, and $\mu$ and $\eta$ are the learning rates associated with the estimated frequency and Fourier series coefficients, respectively. $\hat\omega$ is \textcolor{blue}{used to estimate the person's} cadence in the experiment described in Section \ref{sec:exp}.


\subsection{Learning of the Cue Response Model}\label{subsec:gp}
In order to provide personalized assistance that accounts for the individual's response to the feedback, a solution is formulated based on the HIL framework. A Gaussian process (GP) is used to model the person's response to a given auditory cue while walking at a given cadence. Specifically, 
\begin{align}
&\hat\omega_k = \hat\omega_t \label{incr}\\
&Y=f(X)+H\beta, \\
&\mbox{where } f(X)\sptilde GP(m(X), k(X,X')) \notag\\
&Y = \hat\omega_k, X = (\hat\omega_{k-1},c_{k-1}) \notag
\end{align}

\noindent where $\hat\omega_k$ is the estimated cadence at time \textit{t} from the CDS model; the index, \textit{k}, increments every four strides; \textcolor{blue}{when $k$ is incremented at time $t$, $\hat\omega_t$ is} directly written to $\hat\omega_k$ as shown in Eq~\ref{incr}; $c_k$ denotes the auditory cue frequency provided at increment $k$ and is zero when no cue is provided. Both $\hat\omega$ and \textit{c} are in Hertz (Hz). The GP prior, \textit{f(X)}, is computed over the available data up to index \textit{k}, where $Y$ is a list of the cadences, and $X$ is a list of the preceding cadences and cue frequencies. New data gets appended to \textit{X} and \textit{Y} with each $k$. \textit{m(X)} is the mean function and \textit{k(X,X')} is the square exponential kernel of the GP. An explicit, constant basis function, $H$, is specified, where $H$ is a k-by-one vector of ones and $\beta$ is a scalar basis coefficient estimated from the data.

\textcolor{blue}{The approach is similar to the Bayesian approach described in \cite{kim2017human}. However, an initial exploration with a pre-defined set of parameters was not performed. Instead, initialization was done through random exploration until sufficient data is collected to compute the gradient, since the action space is small and the GP only requires a small number of samples.}

\subsection{Cue Provision and Optimization}\label{subsec:opt}

During the GP update, we also check whether the participant's current cadence is within a threshold of the target cadence ($w_{target}$). If $\mid\hat\omega_t-w_{target}\mid > threshold$, the expected value of the predictive posterior distribution is computed using the GP model given the current cadence and the available range of cue frequencies ($\pm35$\% of the baseline cadence, $\omega_{baseline}$) to minimize the difference between the mean and the target cadence, as follows: 
\begin{align}\label{eq:minimize}
&\hat{\overline{\omega}}_{k+1} = k((\hat\omega_{k},c_{k}), X)(k(X,X)+\sigma^2 I_n)^{-1}(Y-H\beta) \notag \\
&\mathbf{J}(\hat\omega_{k},c_{k}) = (\omega_{target}-\hat{\overline{\omega}}_{k+1})^2 \notag \\
&c_{k} = \argmin_{c_{k}} \mathbf{J}(\hat\omega_{k},c_{k}), \mbox{ subject to } \notag \\
&\omega_{baseline}\times0.65 \leq c_{k} \leq \omega_{baseline}\times1.35
\end{align}
\noindent where $\hat{\overline{\omega}}_{k+1}$ is the next cadence estimated from the GP model, $I_n$ is a square identity matrix. The minimization algorithm is initialized with a randomly generated number. This random initialization is used for response space exploration; when a random starting point far away from the kernel is selected, the optimizer will exit immediately as the size of the gradient is less than the optimality tolerance. The random selection behaviour will stop once the gradient can be computed. Based on this property, the model can be interpreted as having two phases: the exploration (exp) phase (i.e. random sampling) versus the converged (cvg) phase (i.e. when there is a valid gradient). The two phases are discussed in Section \ref{sec:discussion}.

\begin{figure}[htb]
\begin{center}
\includegraphics[scale=0.7]{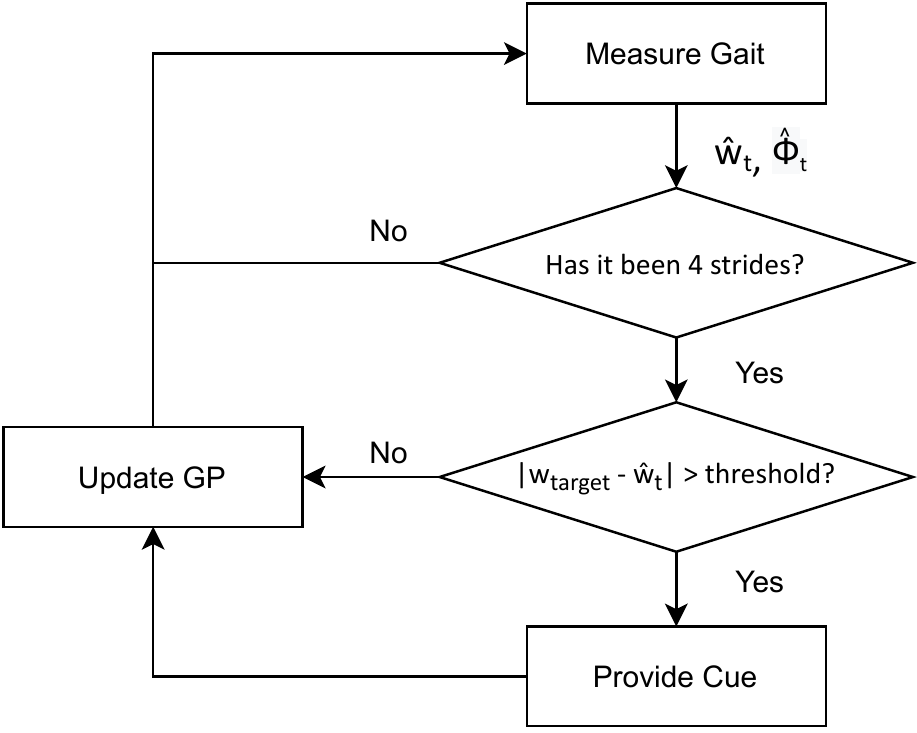}
\caption{With every new CDS estimate, \textcolor{blue}{the algorithm utilizes the phase wraparound to track the number of strides. The algorithm checks whether the person's cadence is within the threshold every 4 strides. The system computes and outputs the optimal cue if the threshold is exceeded} and updates the response model.}
\label{fig:Code}
\end{center}
\end{figure}

The personalized cue provision algorithm described above is summarized in Figure \ref{fig:Code}. The algorithm is implemented in MATLAB, using the GP models (Statistics and Machine Learning Toolbox) and nonlinear least-squares optimization using trust region methods (Optimization Toolbox) \cite{MATLAB:2019}.

\section{Experiments}\label{sec:exp}
We examined the effect of different auditory cue-provision strategies on cadence in the experiment, as auditory cues have been shown to have a strong influence on cadence \cite{sweeney2019technological}. 

\subsection{Participants} 
A convenience sample of 25 participants (5 female/20 male; age 26.08$\pm$3.58 years; height 174.52$\pm$8.65 cm; mass 70.88$\pm$12.84 kg; mean$\pm$standard deviation) enrolled in the study. All participants provided consent \textcolor{blue}{at} the start of the experiment. The study (Project ID 22556) was approved by the Monash University Human Research Ethics Committee.

\subsection{\textcolor{blue}{Equipment and Parameter Initialization}}\label{sec:materials}
The motion data was recorded using a single IMU sensor with the WaveTrack Inertial System (Cometa Systems, Milan, IT).  \textcolor{blue}{The equipment is shown in Figure~\ref{fig:exp-diagram}B.} The data was sampled at 285 Hz and streamed wirelessly into a custom program in C\#. The C\# program ran on a laptop (Windows 10, i7 core with no GPU), which controlled the timing of the auditory cues played from the computer and interfaced with MATLAB. The coefficients of the gait parameter estimation algorithm, CDS, were initialized as follows: M = 7, $\mu$ = 0.1, $\eta$ = 1, $\phi_0$ = 0, $\omega_0$ = $2\pi\cdot\frac{4}{5}$, $\alpha_{m,0}$ = 0 for all $\alpha_m$, and $\beta_{m,0}$ = 0 for all $\beta_m$. \textcolor{blue}{The initial values for $\phi_0$, $\alpha_{m}$, and $\beta_{m}$ were set to 0 as there is no strong prior, whereas $\omega_0$ was based on the typical walking speed for the healthy population \cite{waugh2019online}. The parameters were the same as \cite{waugh2019online}, except for M, $\mu$, and $\eta$. Specifically, M was reduced as gyroscope data contains less high frequency content. $\mu$, $\eta$ were manually tuned until the frequency converged within four strides while minimizing oscillations around the settled value. }

\subsection{Experimental Conditions}
There was 1 control and 6 levels in the study. In the control condition, the participants walked at their natural cadence with no cueing. The baseline cadence of each participant ($\omega_{baseline}$) was measured during control and was used to calculate the two target cadences ($\pm20\% \omega_{baseline}$). 

\begin{figure*}[htb]
\begin{center}
\includegraphics[width=\textwidth]{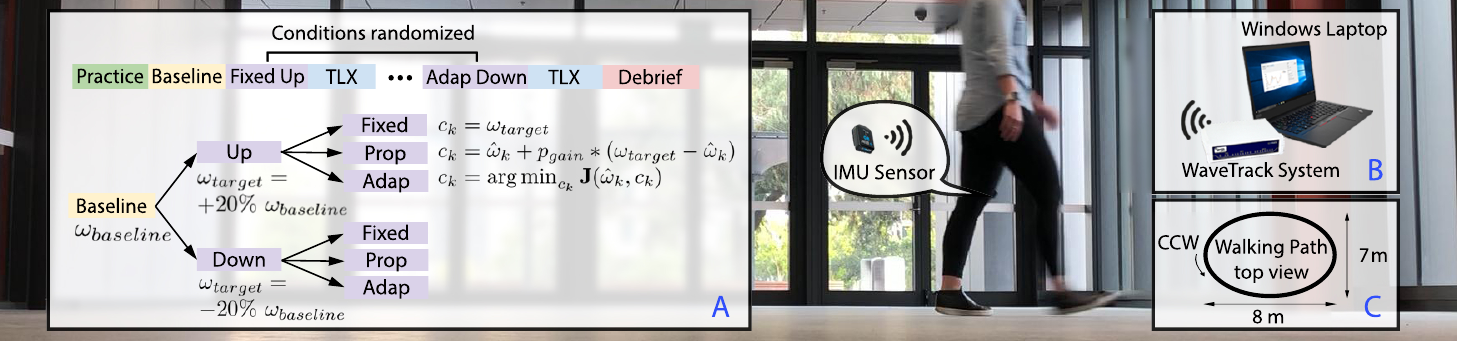}
\caption{\textcolor{blue}{Experimental Procedure:  Panel A illustrates an example experiment sequence and the cueing approaches described in Section~\ref{sec:exp_protocol}. Panel B, with the IMU, illustrates the experiment equipment. Panel C shows the walking space and route. The background image was taken during the experiment.}}
\label{fig:exp-diagram}
\end{center}
\end{figure*}

Following the control condition, each of the three cueing approaches was implemented for each target cadence: fixed, proportional, and adaptive. In the fixed cue approach, beats were provided directly at the target cadence, emulating the baseline cueing mechanisms in the literature \cite{Bachlin2010, Mikos2019}. In the proportional cue approach, the pace of the cue was proportional to the error between the participant's current cadence and the target cadence. The proportional approach serves as an intermediate comparison between the fixed and adaptive approach\textcolor{blue}{, similar to the approach in \cite{zhang2020wearable}. The proportional approach} accounts for the person's current cadence but the error gain requires manual tuning and the gain remains the same throughout the experiment. The proportional gain (i.e. $p_{gain}$) was chosen to be 0.5, which was set empirically during pilot tests. Since the gain was small, the pace of the provided cues was close to the person's current cadence. Finally, the adaptive approach was the algorithm that incorporated the participant's individualized cue response model and optimization, described in Section~\ref{sec:propApp}. \textcolor{blue}{The experiment conditions are summarized in Figure~\ref{fig:exp-diagram}A.}

All three approaches provided cues only when the participant's cadence was out of the acceptable boundary, set to $\pm$ 1\% of the target cadence, as described in Figure \ref{fig:Code}. Eight metronome beats were provided if the acceptable condition was not met, one for each step. The number of beats was selected empirically as observed in the pilot study, where participants were able to change their gait within eight beats and the CDS model was able to converge to the new pace. Each experimental condition took 7 minutes to complete. The duration was as an extension of the Six-Minute-Walk clinical test. During each 7-minute session, cues were provided based on the corresponding condition in the first 6 minutes, and no cue was provided in the last minute. 

\subsection{Experimental Protocol}\label{sec:exp_protocol}
The participant watched an introduction video and placed the IMU sensor above the knee of the dominant leg during preparation. A short training session ($<$1 minute) was provided to allow the participant to become familiar with the act of syncing one’s gait to the metronome beats, where a metronome beat at 1.3 Hz was played continuously for the participant to follow. After training, the participant completed the control condition where they walked in a big circle for 7 minutes without cues. They were told to walk naturally and forget about the practice metronome beats. The participant completed a demographic survey and proceeded to the experimental conditions. The order of the conditions was randomly generated for each participant \textcolor{blue}{and the participants were blinded to the conditions. Each condition was followed by a NASA Task Load Index (TLX) survey.} Finally, the experiment was concluded after a debriefing session.

\subsection{Analysis}
The convergence of the GP was first assessed to validate its ability to model the person's response to cues. The relationship between the experimental conditions and the resultant gait changes was then analyzed using linear mixed-effect models (LME) in R \cite{citeR}. Square root transformation was applied to all data except the task load index score for the following analysis. The transformation helped with the normality and homoscedasticity assumptions during visual inspection of the residual plots. In general, the model satisfies these assumptions but contains outliers towards the tails. Data shown in the box plots in Section \ref{sec:results} are the un-transformed data for an easier interpretation. The fixed effects of the LME model are the different cue-providing approaches and the random effects are the intercepts for the individual participants. P-values were calculated using the likelihood ratio tests between the model without the fixed effect and the model with the fixed and random effects. 

The performance of the cueing approaches (proportional and adaptive) was benchmarked against the baseline fixed cue approach. The analysis was grouped into the speeding up (UP) and slowing down (DOWN) conditions. \textcolor{blue}{The adaptive approach was further divided} based on the two phases of the GP model: the initial exploration (exp) phase during the first 70 seconds of the experiment and converged (cvg) phase for the remaining time in the experiment. \textcolor{blue}{Figure~\ref{present_data} illustrates the exploration and converged phases.  The first three cues in the Adaptive-Down panel are exploratory, as the goal is to slow down but cues that are faster than the target were provided. Afterwards, the cues were consistently around the target cadence, indicating the start of the converged phase. Based on visual inspection of all the data, 70 seconds was chosen as the upper bound for the exploration phase.} 

\section{Results} \label{sec:results}
\begin{figure}[!t]
\centering
\includegraphics[width=3.2in]{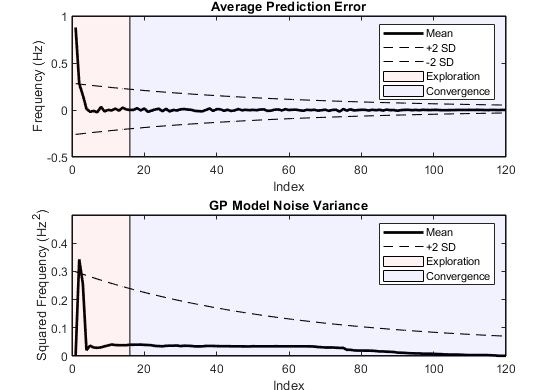}
\caption{Evolution of the GP prediction error and the model error variance averaged over all participants for the adaptive cueing trials. The x-axis is the index value which increments every 4 cycles (i.e. when the acceptable condition is checked). The exploration v.s. the convergence phase is displayed using the average index from all participants.}
\label{gp_performance}
\end{figure}

\noindent\textbf{Adaptive Framework: Response Model Convergence}
The GP modelling error and convergence, averaged over all participants and UP/DOWN conditions, are shown in Figure \ref{gp_performance}. \textcolor{blue}{Both the variance and prediction error are high} during exploration. The variance quickly drops off within 5 iterations as the algorithm learns the response model. However, the error variance\textcolor{blue}{, which indicates the model's confidence,} does not decrease further until later in the experiment due to the fact that similar cues are often provided after the initial exploration. The result shows that GP \textcolor{blue}{can} capture the participant's behaviour around the target cadence.

\noindent\textbf{Sample Experimental Data} A sample dataset from a participant is shown in Figure \ref{present_data}. The following metrics were used to quantify the performance of the cueing approaches: 
\begin{figure}[!t]
\centering
\includegraphics[width=3.4in]{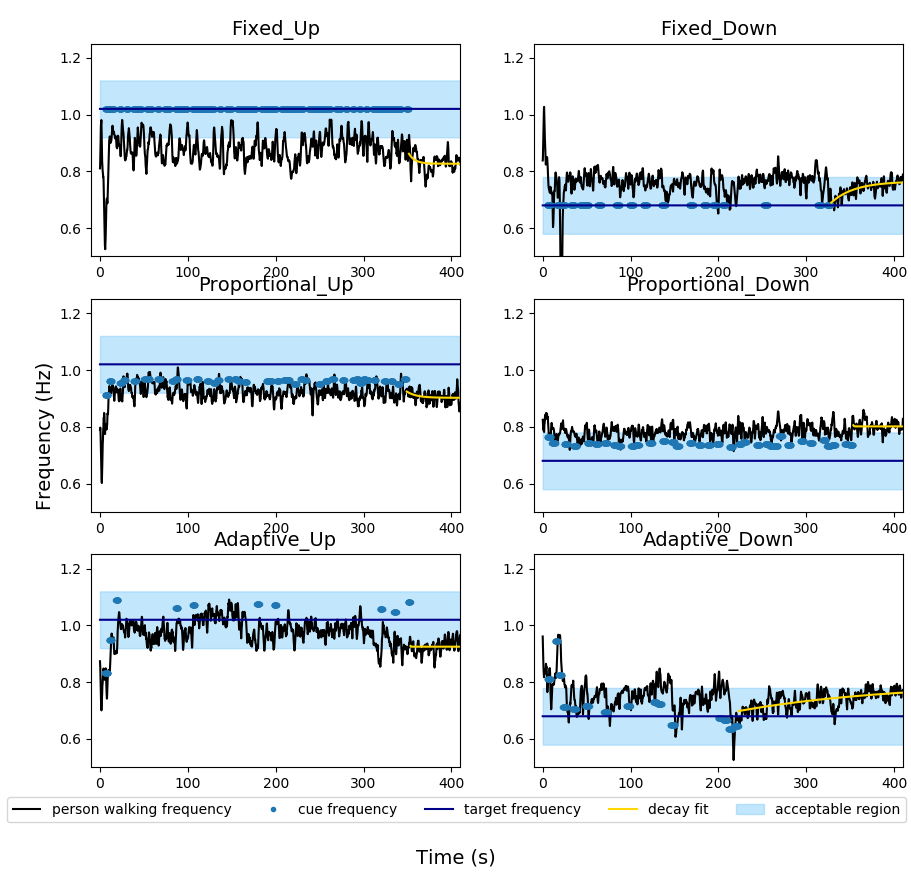}
\caption{Sample data from a participant who had a strong tendency to return to the baseline. Both the fixed and proportional approach attempt to change the cadence by providing more cues, whereas the adaptive approach provides cues at a faster/slower frequency than the target frequency. \textcolor{blue}{The adaptive approach was able to shift the participant's cadence to the desired boundary more effectively compared to other approaches.}}
\label{present_data}
\end{figure}

\begin{figure}[!t]
\centering
\includegraphics[width=3.4in]{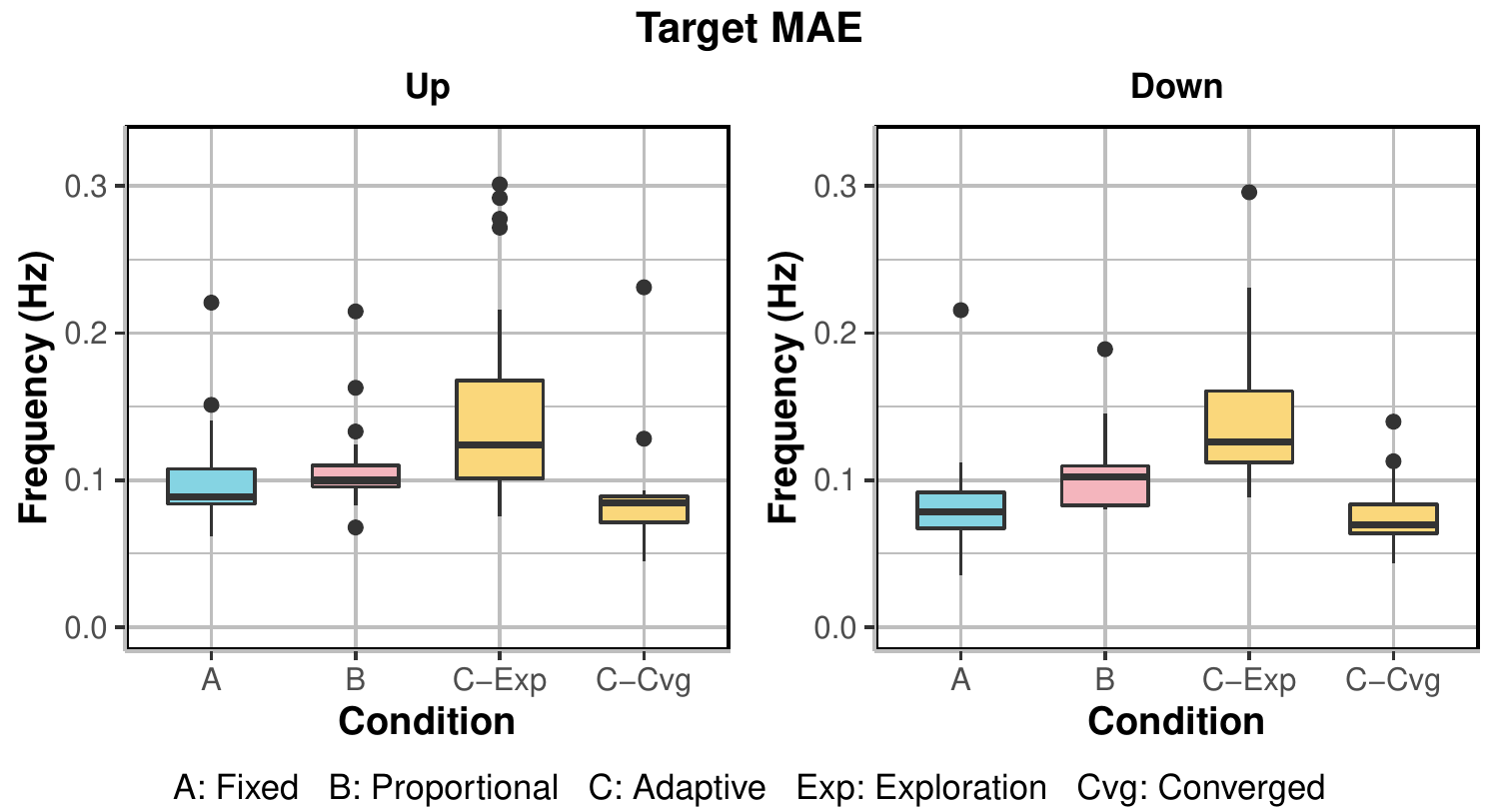}
\caption{The target MAE grouped by the UP (left) and DOWN (right) trials. The metric is further divided for the adaptive approach into two stages based on the GP model convergence: exploration (exp) and converged model (cvg). The \textcolor{blue}{converged} adaptive approach has the lowest target MAE.}
\label{fig:target_mae}
\end{figure}

\noindent\textbf{Target mean absolute error (MAE)}
The target MAE \textcolor{blue}{(Figure \ref{fig:target_mae})} is calculated as the mean absolute error between the participant's estimated cadence and the target cadence. A low target MAE means the participant is able to change their original cadence to match the new target cadence. The LME model for UP shows that the effect of cueing method is significant (likelihood-ratio test statistic ($\lambda_{LR}$) = 25.8837, p $<<$ 0.05, standard deviation of the random effect (StdDev R.N.) = 0.0164). On average, the proportional approach has a higher target MAE compared to the fixed approach (Value = 0.0159, 95\% Confidence Interval (CI) = [-0.0147, 0.0338], Standard Error (SE) = 0.016). The adaptive approach during exploration also has a higher target MAE than the fixed approach (Value = 0.0671, CI = [0.0266, 0.0756], SE = 0.0162). The \textcolor{blue}{converged} adaptive approach has a lower target MAE than the fixed approach (Value = -0.0216, CI = [-0.038, 0.0137], SE = 0.017). 

For the DOWN conditions, the effect of the cueing method is also significant ($\lambda_{LR}$ = 56.3132, p $<<$ 0.05, StdDev R.N. = 0.0276). The target MAE is higher in the proportional approach than the fixed approach (Value = 0.0383, CI = [0.0135, 0.0631], SE = 0.0127) and it is also higher for the adaptive approach during exploration compared to the fixed approach (Value = 0.0975, CI = [0.0724, 0.1227], SE = 0.0129). The \textcolor{blue}{converged }adaptive approach has a lower target MAE compared to the fixed approach (Value = -0.0076, CI = [-0.0338, 0.0185], SE = 0.0134). 

\noindent\textbf{Intermediate MAE and Decay Rate}
Intermediate MAE and decay rate indicate how well the participant is able to maintain the target cadence in the absence of the cue. Intermediate MAE measures the mean absolute error between the participant's current cadence and the target cadence in the periods of silence during the first 6 minutes of the experiment. Decay rate is the rate at which the participant returns to a new steady state cadence after the final cue is provided. We calculate decay rate by fitting an exponential function to the cadence estimate. An example of the fitted decay rate data can be seen in in Figure \ref{present_data}. A low intermediate MAE and a low decay rate would indicate a better maintenance of the new cadence. 

\begin{figure}[!t]
\centering
\includegraphics[width=3.4in]{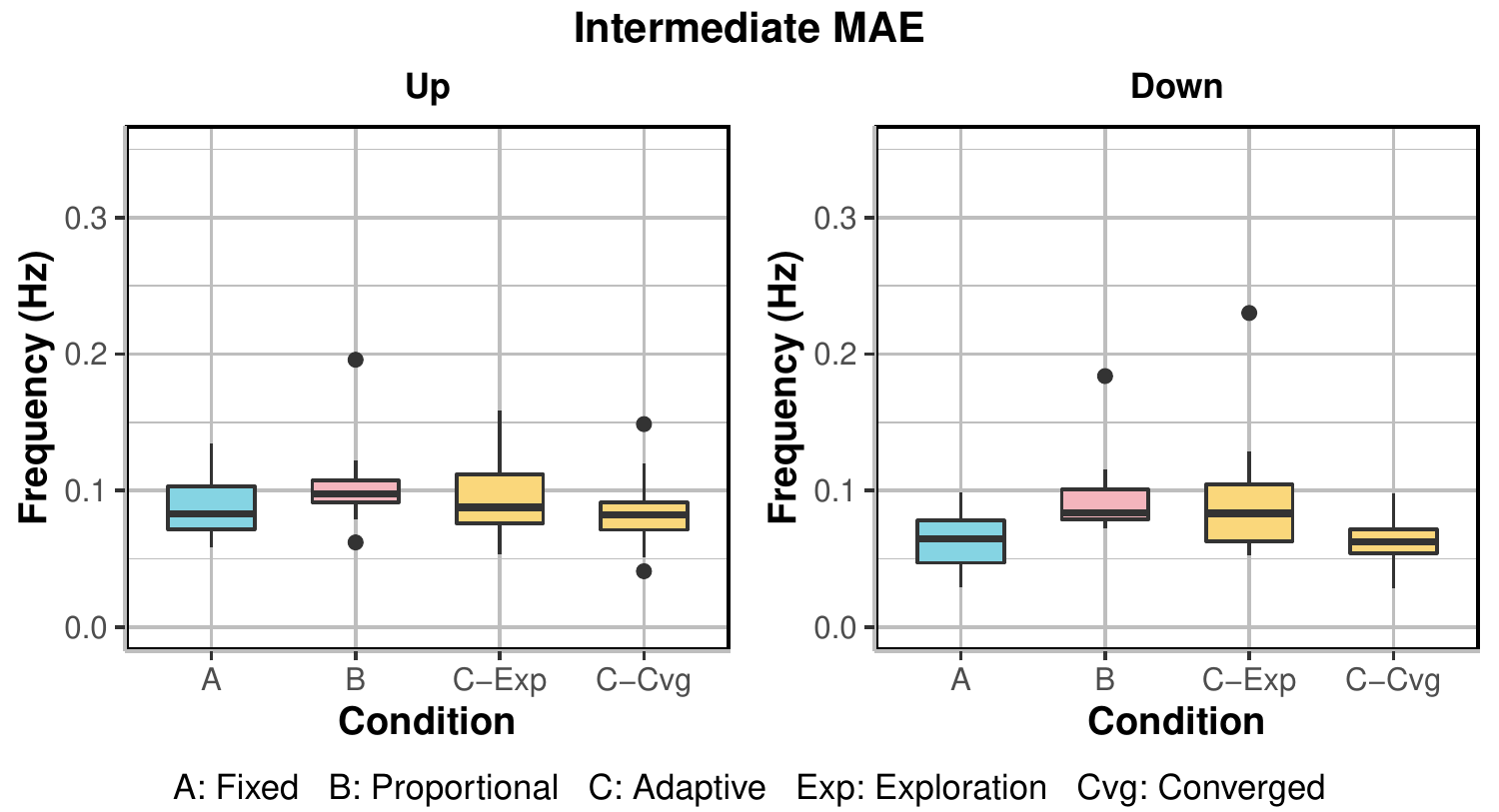}
\caption{The intermediate MAE grouped by the UP (left) and DOWN (right) trials. The intermediate MAE is the highest for the proportional approach in both UP and DOWN trials. However, the effect of cueing conditions is only significant for the DOWN condition.}
\label{fig:intermediate}
\end{figure}

\begin{figure}[!t]
\centering
\includegraphics[width=3.4in]{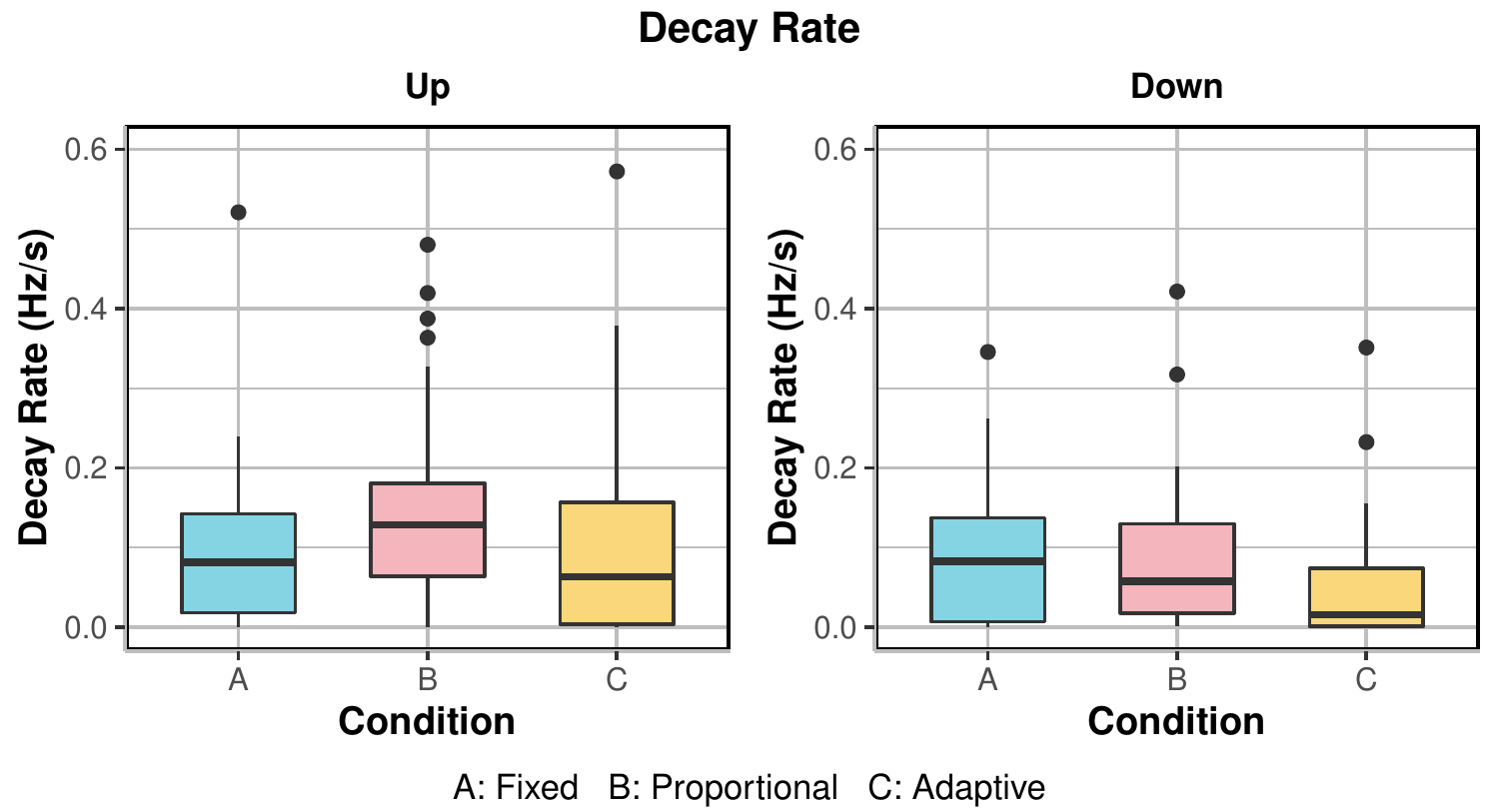}
\caption{The decay rate grouped by the UP (left) and DOWN (right) trials. The effect of the cueing condition is not significant for both trials.}
\label{fig:decay}
\end{figure}

The effect of the cueing method is not significant \textcolor{blue}{for the UP conditions} for the intermediate MAE outcome ($\lambda_{LR}$ = 5.9734, p = 0.1129 $>$ 0.05, StdDev R.N. = 0.0115). For the DOWN conditions, the effect of the cueing approach is significant ($\lambda_{LR}$ = 35.9315, p $<<$ 0.05, StdDev R.N. = 0.0235). The proportional approach has a higher intermediate MAE compared to the fixed approach (Value = 0.0512, CI = [0.0302, 0.0721], SE = 0.0107); the adaptive approach during exploration is also higher than the fixed approach (Value = 0.0434, CI = [0.0217, 0.0653], SE = 0.0111). The converged adaptive approach has an intermediate MAE lower than the fixed cue condition (Value = -0.007, CI = [-0.0285, 0.0144], SE = 0.011). The results are shown in Figure \ref{fig:intermediate}.

Cueing approaches do not significantly affect the decay rate in the UP conditions ($\lambda_{LR}$ = 5.3566, p = 0.0687 $>$ 0.05, StdDev R.N. = 0.1505). Similarly, the cueing approaches also do not significantly affect the decay rate for the DOWN conditions ($\lambda_{LR}$ = 3.8148, p = 0.1485 $>$ 0.05, StdDev R.N. = 0.0756). The results are shown in Figure \ref{fig:decay}.

\noindent\textbf{Percent On}
In terms of minimizing the cue duration to reduce habituation, we quantified the cueing strategy performance using the percent on metric. Percent on represents the amount of time a strategy is providing beats expressed as a percentage of the first 6 minutes of the experiment. 

\begin{figure}[!t]
\centering
\includegraphics[width=3.4in]{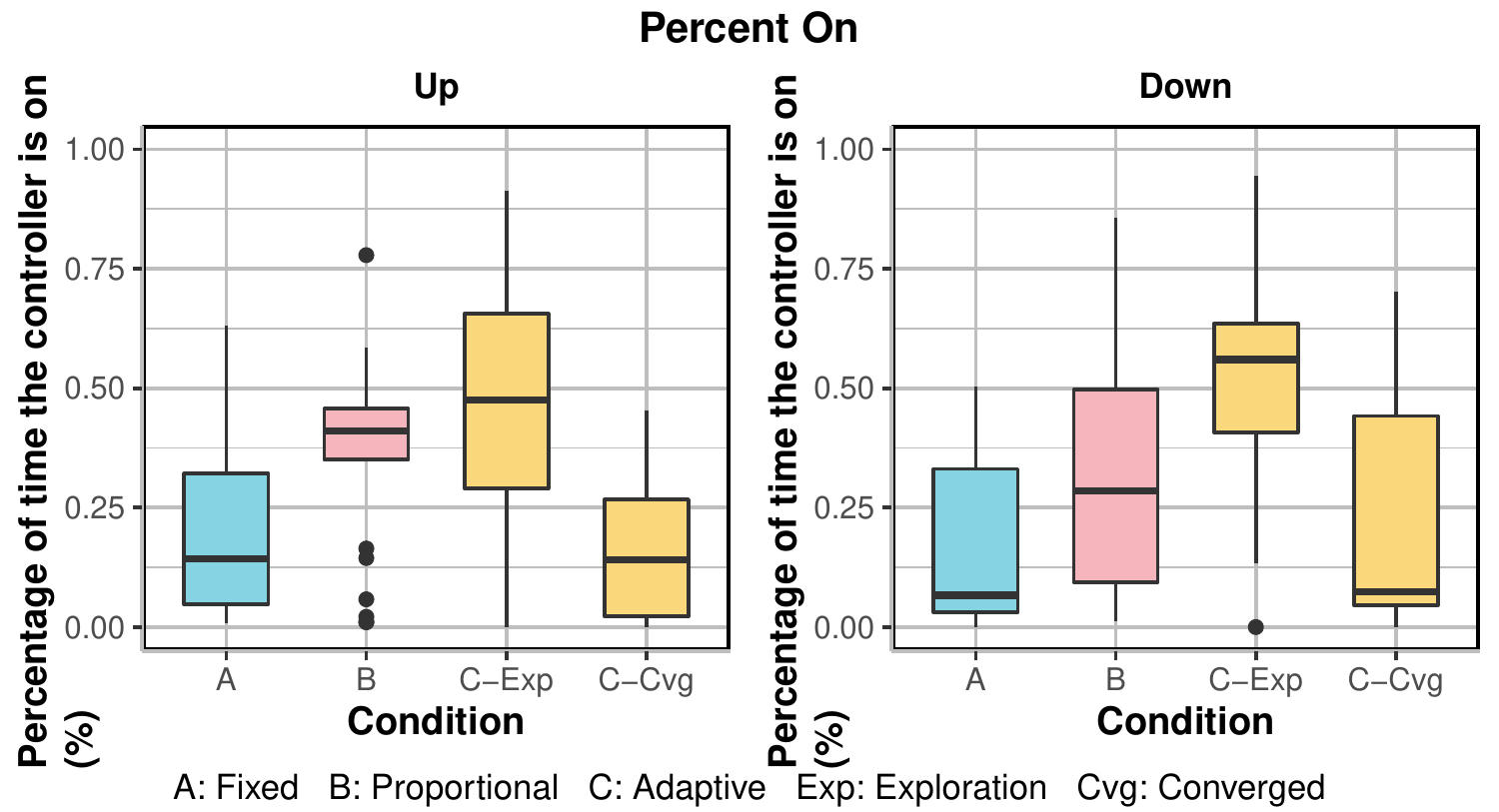}
\caption{The percent on time grouped by the UP (left) and DOWN (right) trials. The percent on time is the highest during the exploration phase of the adaptive approach. \textcolor{blue}{The percent on on time of the converged adaptive approach is close to the fixed approach.}}
\label{fig:percent_on}
\end{figure}

For the UP conditions, the effect of the cueing method is significant. ($\lambda_{LR}$ = 43.2037, p $<<$ 0.05, StdDev R.N. = 0.141). The percent on time is higher for the proportional approach (Value = 0.1625, CI = [0.0679, 0.257], SE = 0.0484); the adaptive approach during exploration also has a higher percent on time than the fixed approach (Value = 0.2507, CI = [0.1562, 0.3453], SE = 0.0484). Once the adaptive approach has converged, the percent on time is lower than the fixed approach (Value = -0.0731, CI = [-0.1676, 0.0214], SE = 0.0484).

For the DOWN conditions, the effect of the cueing approach is also significant ($\lambda_{LR}$ = 44.7463, p $<<$ 0.05, StdDev R.N. = 0.1644). The proportional approach playing cues more than the fixed approach (Value = 0.1517, CI = [0.0596, 0.2436], SE = 0.0471). The adaptive approach during exploration has the highest percent on time compared to the fixed approach (Value = 0.3164, CI = [0.2244, 0.4085], SE = 0.0471). The adaptive approach when converged is also higher than the fixed approach (Value = 0.014, CI = [-0.0779, 0.1061], SE = 0.0471). \textcolor{blue}{The results are displayed in Figure~\ref{fig:percent_on}}.

\begin{figure}[!t]
\centering
\includegraphics[width=3.4in]{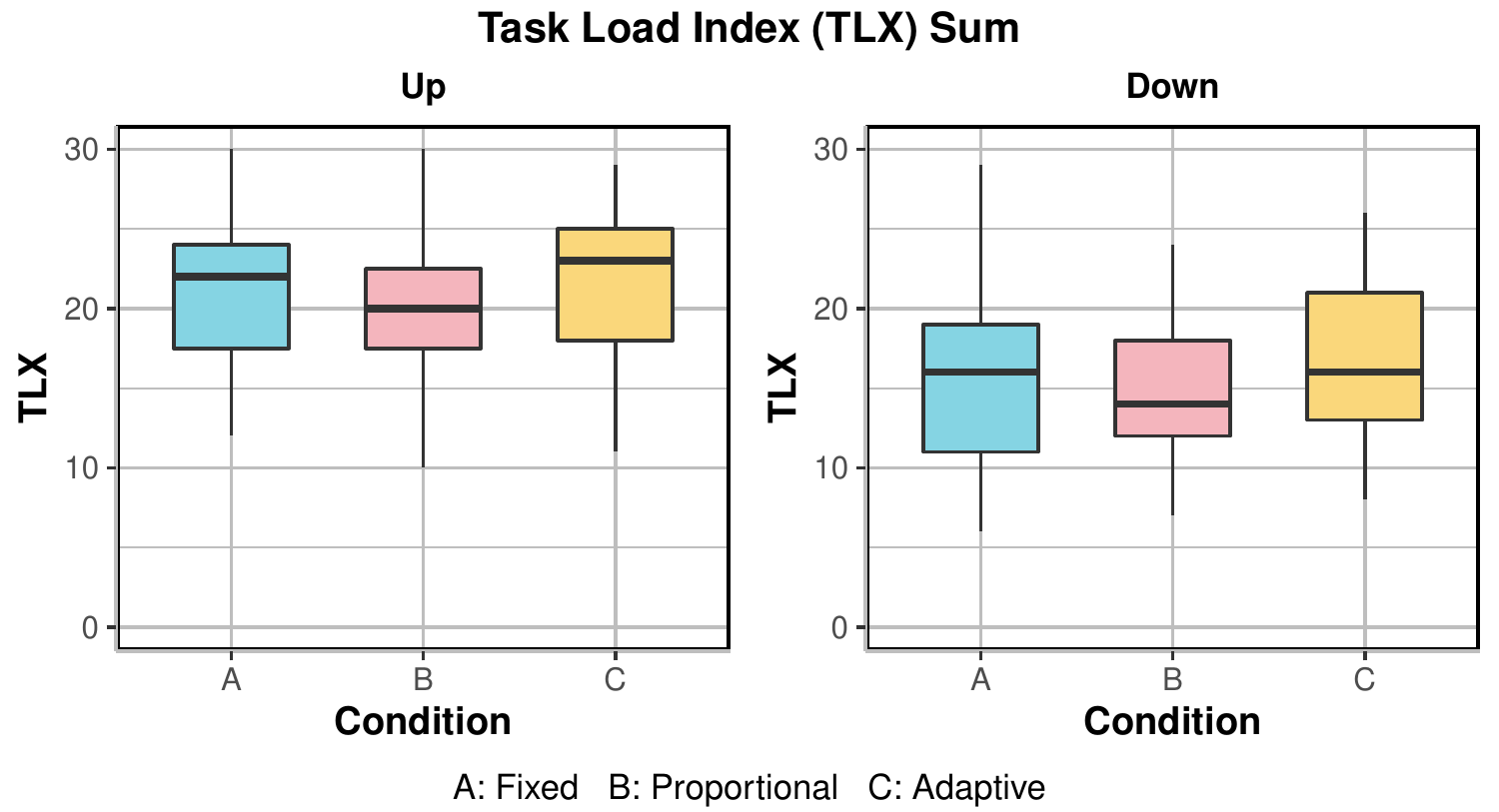}
\caption{The sum of TLX scores across conditions. Overall, the UP conditions are more demanding than the DOWN conditions.}
\label{fig:tlx_raw}
\end{figure}

\noindent\textbf{Participant Perception: NASA-Task Load Index (TLX)}
The sum of the raw TLX scores (\textcolor{blue}{Figure \ref{fig:tlx_raw}}) represents the participant's cognitive workload in each condition. For the UP conditions, the TLX score is not significantly influenced by the cueing approaches ($\lambda_{LR}$ = 1.2837, p = 0.5263 $>$ 0.05, StdDev R.N. = 4.7480). Similarly, the TLX score is not significantly affected by the cueing approaches for the DOWN conditions ($\lambda_{LR}$ = 5.0589, p = 0.0797 $>$ 0.05, StdDev R.N. = 3.4140). 

\section{Discussion}\label{sec:discussion}

\begin{figure}[!t]
\centering
\includegraphics[width=3.4in]{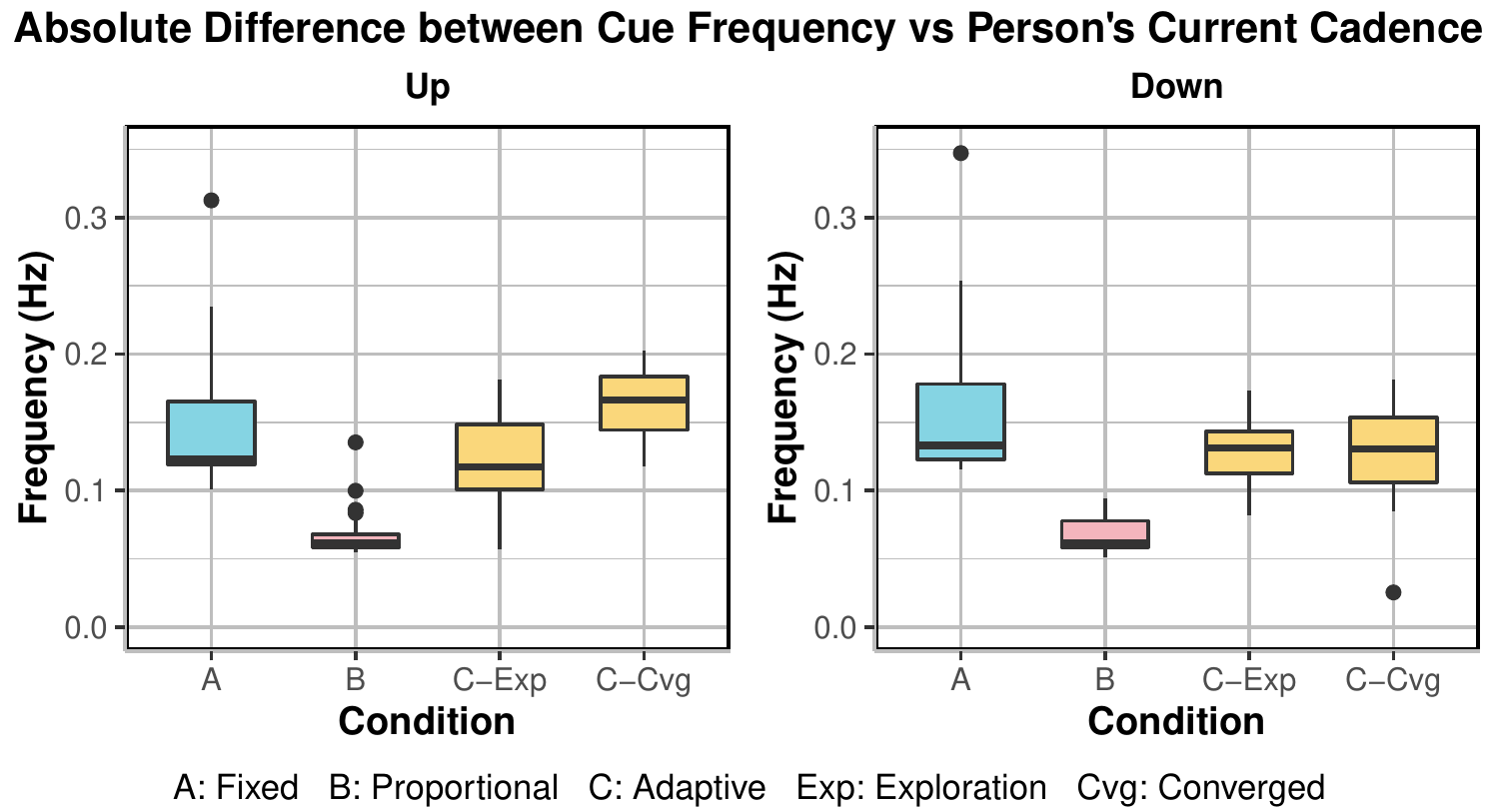}
\caption{The absolute difference gives insights into how the cues are changing the person's cadence. In the UP conditions, the adaptive approach once converged is providing cues at a much higher pace from the participant's current cadence. The proportional approach on the other hand provides participants with cues that are close to their current cadence.}
\label{cue_vs_curr_state}
\end{figure}

\textcolor{blue}{Comparing the different cueing approaches, the adaptive approach is the most effective in achieving the new target cadence. Specifically, the adaptive approach in the converged phase has the lowest target MAE for both target speeds.} The converged adaptive approach reduces the target MAE by providing cues at a very different pace compared to the participant's current state, prompting the participants to be more proactive, as seen in Figure \ref{cue_vs_curr_state}. \textcolor{blue}{Participants also perceived the proactive cueing,  reflected in the higher TLX score. Another factor for the high TLX score is caused by the participants having to follow a random set of beats during the exploration phase. Once GP has converged, the adaptive approach is able to achieve a lower target MAE while having a comparable percent on time to the baseline fixed approach. The adaptive approach might be able to outperform the baseline approach in the future by penalizing cue-playing in the cost function to reduce habituation.} 

\textcolor{blue}{Of the three approaches, the proportional approach is the least effective in changing the natural cadence as seen in the high target MAE and intermediate MAE.} This might be because the proportional cue is not prompting the person to change much from the original cadence. In Figure \ref{cue_vs_curr_state}, it can be seen that the difference between the cue and the current gait cadence is always the smallest for the proportional approach. The phenomenon is due to the choice of the controller gain, which is designed to provide gradual changes in the pace of the cue. The gradual change allows for a lower cognitive workload (as seen in the TLX scores), but is less effective in altering the cadence. While the proportional cue might have been more effective with a higher gain or personalized gain tuning, the cueing method highlights the difficulty in manual gain selection. 

\textcolor{blue}{The three approaches are not significantly different in terms of intermediate MAE and decay rate. This might be due to the decay rate and the intermediate MAE being} influenced by the participant’s memory (i.e. forgetting the pace of the cue over time) and their ability or willingness to adjust their cadence. In the post-study interviews, participants reflected that it was difficult to recall the pace of the cue in the long periods of silence. We also observed patterns in the experiment similar to the participant in Figure \ref{present_data}, where some participants immediately deviate from the target cadence when the cues are off, causing the drastic fluctuations and the on-off cueing pattern.

\textcolor{blue}{Our study has several limitations. Only healthy participants were tested in a single trial. We also assumed that the participant could attain the target, which might not be possible with the patient population. Finally, only auditory cueing was used, which may not be effective with all users and in all environments.}

\textcolor{blue}{The proposed approach could be adopted to provide assistance using exoskeletons. The current HIL approaches (e.g. \cite{kim2017human, zhang2017human, felt2015body}) utilize respiration measurements in the optimization step, which can be difficult to obtain in an everyday setting. With the proposed adaptive framework, kinematics that could replace the respiration measurements, improving usability.}

\section{Conclusions and Future Work}
We proposed an adaptive cueing framework that can simultaneously monitor gait performance of a person and adjust the auditory cues based on the person’s response. In the framework, a Gaussian Process \textcolor{blue}{was used to model} the person’s gait as a function of the provided cues and past gait performance. Using \textcolor{blue}{GP}, personalized assistance can be provided \textcolor{blue}{through optimization} to improve gait performance. We investigated the effectiveness of the adaptive cueing strategy with healthy participants in a gait study, where \textcolor{blue}{the aim was} to change the participant's cadence with the cues. The adaptive cue method \textcolor{blue}{was compared} to the fixed and the proportional methods. The results show that the proportional cues perform the worst among the three cueing approaches, highlighting the need for individualization and adaptation. The adaptive strategy outperforms the both comparison strategies when the GP model has converged. 


\textcolor{blue}{Future work involves developing a wearable device that can provide multi-modal cues for patients with gait impairments. To relax the assumption that participants will be able to achieve to a fixed target gait state, we will adapt the target gait state in real time, and include additional objectives within the cost function to support the provision of multi-modal cues. An advantage of the proposed approach is that the system continuously learns the user's gait profile and response model, and can therefore be used when the patient's response profile is changing (e.g. due to medication or fatigue). We plan to recruit patients to examine the effectiveness of the adaptive cueing strategy.}





\bibliographystyle{IEEEtran}

\bibliography{IEEEabrv,references}

\end{document}